\documentclass{article}

\usepackage{microtype}
\usepackage{graphicx}
\usepackage{subfigure}
\usepackage{booktabs} 

\usepackage{hyperref}

\usepackage[accepted]{icml2019}

\icmltitlerunning{Nonparametric Bayesian Deep Networks with Local Competition}

\usepackage{amsmath}
\usepackage{amssymb}

\usepackage{url}
\usepackage{multirow}
\newcommand\Tstrut{\rule{0pt}{2.ex}}         
\newcommand\Bstrut{\rule[-1.2ex]{0pt}{0pt}}   

\newcommand\scalemath[2]{\scalebox{#1}{\mbox{\ensuremath{\displaystyle #2}}}}

\makeatletter

\let\SF@@footnote\footnote
\def\footnote{\ifx\protect\@typeset@protect
    \expandafter\SF@@footnote
  \else
    \expandafter\SF@gobble@opt
  \fi
}
\expandafter\def\csname SF@gobble@opt \endcsname{\@ifnextchar[
  \SF@gobble@twobracket
  \@gobble
}
\edef\SF@gobble@opt{\noexpand\protect
  \expandafter\noexpand\csname SF@gobble@opt \endcsname}
\def\SF@gobble@twobracket[#1]#2{}

\usepackage{tikz}
\usetikzlibrary{shapes.misc, positioning}
\usetikzlibrary{backgrounds,fit,matrix, calc}
\tikzset{
   box/.style = {minimum height=10pt, minimum width=10pt, draw, rounded corners,rectangle, fill=white!50},
}
\usetikzlibrary{fadings}

\tikzset{
   boxconv/.style = {minimum height=2cm, minimum width=2cm, draw, line width=0.4mm, fill opacity=0.9, rounded corners,rectangle, fill=white!50},
}

\tikzset{
   boxconv_inactive/.style = {minimum height=2cm, minimum width=2cm,line width=0.3mm, draw, line width=0.1mm , fill opacity=0.9, rounded corners,rectangle, gray, fill=white!50},
}
\tikzset{
   input/.style = {minimum height=3cm, minimum width=3cm, draw, , fill opacity=0.9, rectangle, fill=white!50},
}

\tikzset{
   boxpooled/.style = {minimum height=1.5cm, minimum width=1.5cm, draw, line width=0.4mm, fill opacity=0.9, rounded corners,rectangle, fill=white!50},
}

\tikzset{
   boxpooled_inactive/.style = {minimum height=1.5cm, minimum width=1.5cm, draw, line width=0.4mm, fill opacity=0.9, rounded corners,rectangle, gray, fill=white!50},
}

\usetikzlibrary{fadings}\tikzset{
    boxwta/.style={%
        draw=black, thick,
        rectangle,
        rounded corners,
        minimum height=3cm,
        minimum width=3cm
    }
}

\usetikzlibrary{fadings}\tikzset{
    box1/.style={%
        draw=black, thick,
        rectangle,
        minimum height=2cm,
        minimum width=2cm
    }
}

\usetikzlibrary{fadings}\tikzset{
    box2/.style={%
        draw=black, thick,
        rectangle,
        minimum height=1.cm,
        minimum width=1.cm
    }
}

\usetikzlibrary{fadings}\tikzset{
    box3/.style={%
        draw=black, thick,
        rectangle,
        minimum height=.8cm,
        minimum width=.8cm
    }
}
\usetikzlibrary{decorations.markings}

\begin{document}

\twocolumn[
\icmltitle{Nonparametric Bayesian Deep Networks with Local Competition}



\icmlsetsymbol{equal}{*}

\begin{icmlauthorlist}
\icmlauthor{Konstantinos P. Panousis}{equal,di}
\icmlauthor{Sotirios Chatzis}{equal,goo}
\icmlauthor{Sergios Theodoridis}{di,ch}
\end{icmlauthorlist}

\icmlaffiliation{di}{Dept. of Informatics \& Telecommunications, National and Kapodistrian University of Athens, Greece}
\icmlaffiliation{goo}{Dept. of Electrical Eng., Computer Eng., and Informatics, Cyprus University of Technology, Limassol, Cyprus}
\icmlaffiliation{ch}{The Chinese University of Hong Kong, Shenzen, China}

\icmlcorrespondingauthor{Konstantinos P. Panousis}{kpanousis@di.uoa.gr}
\icmlcorrespondingauthor{Sotirios Chatzis}{sotirios.chatzis@cut.ac.cy}

\icmlkeywords{Machine Learning, ICML}

\vskip 0.3in
]



\printAffiliationsAndNotice{\icmlEqualContribution} 


\begin{abstract}
The aim of this work is to enable inference of deep networks that retain high accuracy for the least possible model complexity, with the latter deduced from the data during inference.  To this end, we revisit deep networks that comprise competing linear units, as opposed to nonlinear units that do not entail any form of (local) competition. In this context, our main technical innovation consists in an inferential setup that 
leverages solid arguments from Bayesian nonparametrics. We infer both the needed set of connections or locally competing sets of units, as well as the required floating-point precision for storing the network parameters. Specifically, we introduce auxiliary discrete latent variables representing which initial network components are actually needed for modeling the data at hand, and perform Bayesian inference over them by imposing appropriate stick-breaking priors.  As we experimentally show using benchmark datasets, our approach yields networks with less computational footprint than the state-of-the-art, and with no compromises in predictive accuracy.
\end{abstract}

\section{Introduction}

Deep neural networks (DNNs)~\cite{lecun2015deep} are flexible models that represent complex functions as a 
combination of simpler primitives. Despite their success in a wide range of applications, they typically 
suffer from overparameterization: they entail millions of weights, a large fraction of which is actually 
redundant. This leads to unnecessary computational burden, and limits their scalability to commodity hardware 
devices, such as mobile phones and cars. In addition, this fact  renders DNNs susceptible to strong overfitting tendencies that may severely undermine their generalization capacity. 

The deep learning community has devoted significant effort to address overfitting in deep learning; $\ell_2$ 
regularization, Dropout, and variational variants thereof are characteristic such examples 
\cite{Gal2015DropoutB}. However, the scope of regularization is limited to  effectively training (and 
retaining) all network weights. Addressing redundancy in deep networks requires data-driven structure shrinkage and weight compression techniques.

A popular type of solution to this end consists in training a condensed student network by leveraging a 
previously trained full-fledged teacher network~\cite{dodeep,Distilling}. However, this paradigm suffers from 
two main drawbacks: (i) One cannot avoid the computational costs and overfitting tendencies related to 
training a large deep network; on the contrary, the total training costs are augmented with the weight 
distillation and training costs of the student network; and (ii) the student teaching procedure itself entails 
a large deal of heuristics and assorted artistry in designing effective teacher distillation. 

As an alternative, several researchers have examined application of network component (unit/connection) 
pruning criteria. In most cases, these criteria are applied on top of some appropriate regularization 
technique. In this context, Bayesian Neural Networks (BNNs) have been proposed as a full probabilistic 
paradigm for formulating DNNs~\cite{Gal2015DropoutB,Graves2011}, obtained by imposing a prior distribution over their weights. Then, appropriate posteriors are inferred, and predictive 
distributions are obtained via marginalization in the Bayesian averaging sense. This way, BNNs induce strong 
regularization under a solid inferential framework. In addition, they naturally allow for reducing floating-point precision in storing the network weights. Specifically, 
since Bayesian inference boils down to drawing samples from an inferred weight posterior, the higher the 
inferred weight posterior variance, the lower the needed floating-point precision~\cite{Welling}.

Finally,~\citet{Chatzis18} recently considered addressing these problems by introducing an additional set of auxiliary Bernoulli latent variables, which 
explicitly indicate the utility of each component (in an ``on/off'' fashion). In this context, they obtain a sparsity-inducing behavior, by imposing appropriate stick-breaking priors~\cite{Ishwaran2001} over the postulated auxiliary latent variables. Their study, although \emph{limited to variational autoencoders}, showed promising results in a variety of benchmarks.

On the other hand, a prevalent characteristic of modern deep networks is the use of nonlinear units on each 
hidden layer. Even though this sort of functionality offers a mathematically convenient way of creating a 
hierarchical model, it is also well understood that it does not come with strong biological plausibility. 
Indeed, there is an increasing body of evidence supporting that neurons in biological systems that have 
similar functional properties are aggregated together in modules or columns where local competition takes 
place~\cite{principles_neural_science,sveen, eccles, Stefanis, Douglas, lasner}. This is effected via the 
entailed lateral inhibition mechanisms, under which only a single neuron within a block can be active at a 
steady state. 

Drawing from this inspiration, several researchers have examined development of deep networks which replace 
nonlinear units with local competition mechanisms among simpler linear units. As it has been shown, such 
local winner-takes-all (LWTA) networks can discover effective sparsely distributed  representations of their 
input stimuli~\cite{lee99,ols96},  and constitute universal function approximators, as powerful as networks 
with threshold or sigmoidal units~\cite{Maass:1999, Maass:2000}. In addition, this  type of network 
organization has been argued to give rise to a number of interesting properties, including automatic gain 
control, noise suppression, and prevention of catastrophic forgetting in online learning 
\cite{Compete,grossberg,mccloskey}.

This paper draws from these results, and attempts to offer a principled and systematic paradigm for inferring the needed network complexity and compressing its parameters. We posit that the 
capacity to infer an explicit posterior distribution of component (connection/unit) utility in the 
context of LWTA-based deep networks may offer significant advantages in model effectiveness and computational 
efficiency. The proposed inferential construction relies on nonparametric Bayesian inference arguments, namely stick-breaking priors; we employ these tools in a fashion tailored to the unique structural characteristics of LWTA networks. 
This way, we give rise to a data-driven mechanism that intelligently adapts the complexity of model structure and infers the needed floating-point precision. 

We derive efficient training and inference algorithms for our model, by relying 
on stochastic gradient variational Bayes (SGVB).  We dub our approach Stick-Breaking LWTA (SB-LWTA) networks. 
We evaluate our approach using well-known benchmark datasets. Our provided empirical evidence vouches for the 
capacity of our approach to yield predictive accuracy at least competitive with the state-of-the-art, while 
enabling automatic inference of the model complexity, concurrently with model parameters estimation. This results in trained networks that yield much better memory footprint than the competition, without the need of extensively applying heuristic criteria.      

The remainder of this paper is organized as follows: In Section 2, we introduce the proposed approach. In Section 3, we provide the  training and inference algorithms of our model. In Section 4, we perform an extensive experimental evaluation of our approach, and provide insights into its functionality. Finally, in the concluding Section, we summarize the contribution of this work, and discuss directions for further research.

\section{Proposed Approach}
\begin{figure*}
\begin{center}
{\scalebox{.58}{\def\layersep{3.5cm}
\def\inputsize{2}
\def\wtablocks{2}
\def\neuronsep{5}
\def\outputsize{2}
\def\wtasep{2.5}
\def\wtablocks{3}
\def\unitsperblock{2}
\def\prob{0.55}

\begin{tikzpicture}[-,draw, node distance=\layersep, label/.style args={#1#2}{%
    postaction={ decorate,
    decoration={ markings, mark=at position #1 with \node #2;}}}]
    \tikzstyle{every pin edge}=[<-,shorten <=1pt]
    \tikzstyle{neuron}=[circle,draw,minimum size=12pt,inner sep=0pt]
    \tikzstyle{wta} = [rounded corners,rectangle, draw, minimum_height=3cm, minimum_width=2cm]

    \tikzstyle{input neuron}=[neuron, fill=white!50];
    \tikzstyle{output neuron}=[neuron, fill=white];
    \tikzstyle{hidden neuron}=[neuron, fill=white!50];
    \tikzstyle{hidden neuron activated}=[neuron, fill=white!50, very thick];
    \tikzstyle{wta block} =[wta];
    \tikzstyle{annot} = [text width=10em, text centered]

    \node[input neuron, pin=left:$x_1$] (I-1) at (0,-\neuronsep+\wtasep) {};
    \node[input neuron, pin=left:$x_J$] (I-2) at (0,-2*\neuronsep+\wtasep) {};
		
     	\matrix (H-1) at (\layersep, -\wtasep-0.*\wtasep) [row sep=4mm, column sep=2mm, inner sep=3mm, box, matrix of nodes] 
      {
        		\node[hidden neuron activated](o1-1){}; \\
        		\node[hidden neuron](o1-2){}; \\
	  };

	  
	  \matrix (H-2) at (\layersep, -3*\wtasep-0.*\wtasep) [row sep=4mm, column sep=2mm, inner sep=3mm, box, matrix of nodes] 
      {
        		\node[hidden neuron activated](o2-1){}; \\
        		\node[hidden neuron](o2-2){}; \\
	  };

	\matrix (H2-1) at (2*\layersep, -\wtasep-0.*\wtasep) [row sep=4mm, column sep=2mm, inner sep=3mm, box, matrix of nodes] 
      {
        		\node[hidden neuron](o21-1){}; \\
        		\node[hidden neuron activated](o21-2){}; \\
	  };

	  \matrix (H2-2) at (2*\layersep, -3*\wtasep-0.*\wtasep) [row sep=4mm, column sep=2mm, inner sep=3mm, box, matrix of nodes] 
      {
        		\node[hidden neuron](o22-1){}; \\
        		\node[hidden neuron activated](o22-2){}; \\
	  };
	  
        
    \foreach \name / \y in {1,...,\outputsize}
    		\node[output neuron] (O-\name) at (3*\layersep,-\neuronsep*\y+\wtasep) {};

    \path (I-1) -- (I-2) node [black, font=\Huge, midway, sloped] {$\dots$};
 	\path[black!100, line width=0.55pt, thick] (I-1.east) edge node[ above, rotate=5, black!100] {\scriptsize $z_{1,1}=1$} (o1-1);
    \path[black!100, line width=0.55pt, thick] (I-1.east) edge node[ below, rotate=-15, black!100] {\scriptsize $z_{1,1}=1$} (o1-2.west);
    
    \path[black!100, line width=0.55pt, thick] (I-1.east) edge (o2-1.west);
    \path[black!100, line width=0.55pt, thick] (I-1.east) edge (o2-2.west);

	\path[black!100, line width=0.55pt, thick] (I-2.east) edge (o1-1.west);
    \path[black!100, line width=0.55pt, thick] (I-2.east) edge (o1-2.west);
    
    \path[black!40] (I-2.east) edge node[above, rotate=5, black!100] {\scriptsize $z_{J,K}=0$} (o2-1.west);
    \path[black!40] (I-2.east) edge node[below, rotate=-15, black!100] {\scriptsize $z_{J,K}=0$} (o2-2.west);
    
    
            
    \path (H-1) -- (H-2) node [black, font=\Huge, midway, sloped] {$\dots$};
    \path (H2-1) -- (H2-2) node [black, font=\Huge, midway, sloped] {$\dots$};

    \path[color=black!100, line width=0.55pt, thick] (o1-1.east) edge (o21-1.west); 
    \path[color=black!100, line width=0.55pt, thick] (o1-1.east) edge (o21-2.west);

    \path[color=black!40] (o1-1.east) edge (o22-1.west);  
    \path[color=black!40] (o1-1.east) edge (o22-2.west); 
    
    \path[black!40] (o1-2.east) edge (o21-1.west);  
    \path[black!40] (o1-2.east) edge (o21-2.west); 

    
    \path[black!40] (o1-2.east) edge (o22-1.west);  
    \path[black!40] (o1-2.east) edge (o22-2.west);

    \path[color=black!100, line width=0.55pt, thick] (o2-1.east) edge (o21-1.west);  
    \path[color=black!100, line width=0.55pt, thick] (o2-1.east) edge (o21-2.west);
    
    \path[color=black!100, line width=0.55pt, thick] (o2-1.east) edge (o22-1.west);  
    \path[color=black!100, line width=0.55pt, thick] (o2-1.east) edge (o22-2.west);
    
    \path[black!40] (o2-2.east) edge (o21-1.west);  
    \path[black!40] (o2-2.east) edge (o21-2.west);
    
    \path[black!40] (o2-2.east) edge (o22-1.west);  
    \path[black!40] (o2-2.east) edge (o22-2.west);
    
    \path (O-1) -- (O-2) node [black, font=\Huge, midway, sloped] {$\dots$};

    
    
    

    
    
    

    \path[black!40] (o21-1.east) edge (O-1.west);  
    \path[black!40] (o21-1.east) edge (O-2.west);
    
    \path[black!100, line width=0.55pt, thick] (o21-2.east) edge (O-1.west);  
    \path[black!100, line width=0.55pt, thick] (o21-2.east) edge (O-2.west);

    \path[black!40] (o22-1.east) edge (O-1.west);  
    \path[black!40] (o22-1.east) edge (O-2.west);
    
    \path[black!100, line width=0.55pt, thick] (o22-2.east) edge (O-1.west);  
    \path[black!100, line width=0.55pt, thick] (o22-2.east) edge (O-2.west);

    

    \node[annot,above= -1mm of o1-1] (k) {\scriptsize $\xi$= 1};
    \node[annot,below= -1mm of o1-2] (k) {\scriptsize $\xi$= 0};
    
    \node[annot,above= -1mm of o22-1] (k) {\scriptsize $\xi$= 0};
    \node[annot,below= -1mm of o22-2] (k) {\scriptsize $\xi$= 1};
    \node[annot,above of=H-1, node distance=1.8cm] (hl) {SB-LWTA layer};
    \node[annot,above left=-3mm and -1.7cm of H-1] (k) {\scriptsize $1$};
    
    \node[annot,above left=-3mm and -1.7cm of H2-1] (k) {\scriptsize $1$};
    
    \node[annot,below left=-3mm and -1.7cm of H-2] (k) {\scriptsize $K$};
    
    \node[annot,below left=-3mm and -1.7cm of H2-2] (k) {\scriptsize $K$};
    
    \node[annot,above of=H2-1, node distance=1.8cm] (hl2) {SB-LWTA layer};
    \node[annot,left of=hl] {Input layer};
    \node[annot,right of=hl2] {Output layer};
\end{tikzpicture}
}}
\end{center}
\caption{A graphical illustration of the proposed architecture. Bold edges denote active (effective) connections (with $z=1$); nodes with bold contours denote winner units (with $\xi=1$); rectangles denote LWTA blocks. We consider $U=2$ competitors in each LWTA block, $k=1,\dots,K$.}
\label{fig:wta1}
\end{figure*}
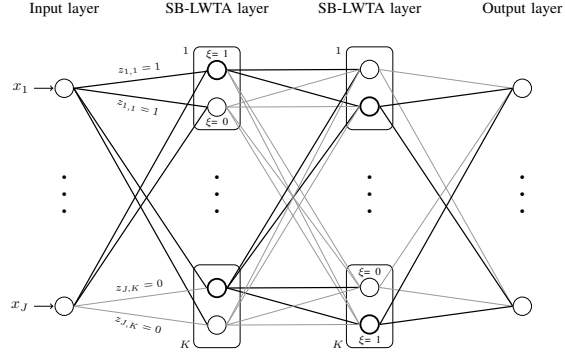

In this work, we introduce a paradigm of designing deep networks whereby
the output of each layer is derived from blocks of competing linear units, and appropriate arguments from nonparametric statistics are employed to infer network component utility in a Bayesian sense. An outline of the envisaged modeling rationale is provided in Fig. \ref{fig:wta1}. 

In the following, we begin our exposition by briefly introducing the Indian Buffet Process (IBP) prior \cite{Griffiths05}; we employ this prior to enable inference of which components introduced into the model at initialization time are actually needed for modeling the data at hand. Then, we proceed to the definition of our proposed model.

\subsection{The Indian Buffet Process}
The IBP is a probability distribution over infinite binary matrices. By using it as a prior, it allows for inferring how many components are needed for modeling a given set of observations, in a way that ensures sparsity in the obtained representations \cite{theo}. In addition, it also allows for the emergence of new components as new observations appear. 
\citet{TehGorGha2007} presented a stick-breaking construction for the IBP, which renders it amenable to variational inference. Let us consider $N$ observations, and denote as $\boldsymbol{Z}=[z_{i,k}]_{i,k=1}^{N,K}$ a binary matrix where each entry indicates the existence of component $k$ in observation $i$. Taking the infinite limit, $K\rightarrow \infty$, we arrive at the following hierarchical representation for the IBP \cite{TehGorGha2007}:
\begin{equation*}
\begin{aligned}[t]
u_k &\sim \text{Beta}(\alpha,1)
\end{aligned}
\quad 
\begin{aligned}[t]
\pi_k = \prod_{i=1}^k u_i \\
\end{aligned}
\quad
\begin{aligned}[t]
z_{ik}&\sim \text{Bernoulli}(\pi_k)\quad  
\end{aligned}
\end{equation*}
Here, $\alpha>0$ is the innovation hyperparameter of the IBP, which controls the magnitude of the induced sparsity. In practice, $K\rightarrow\infty$ denotes a setting whereby we obtain an overcomplete representation of the observed data; that is, $K$ equals input dimensionality.

\subsection{Model Formulation}

Let $\{\boldsymbol{x}_n\}_{n=1}^{N}\in\mathbb{R}^{J}$ be an input dataset containing $N$ observations, with $J$ features each. Hidden layers in traditional neural networks contain nonlinear units; they are presented with linear combinations of the inputs, obtained via a weights matrix $\boldsymbol{W}\in\mathbb{R}^{J\times K}$, and produce output vectors $\{\boldsymbol{y}_n\}_{n=1}^N\in\mathbb{R}^{K}$ as input to the next layer. In our approach, this mechanism is replaced by the introduction of LWTA blocks in the hidden layers, each containing a set of competing linear units. The layer input is originally presented to each block, via different weights for each unit; thus, the weights of the connections are now organized into a three-dimensional matrix $\boldsymbol{W}\in \mathbb{R}^{J\times K \times U}$, where $K$ denotes the number of blocks and $U$ is the number of competing units therein.

Let us consider a layer of the proposed model. Within each block, the linear units compute their activations; then, the block selects one winner unit on the basis of a \emph{competitive random sampling} procedure we describe next, and sets the rest to zero. This way, we yield a \emph{sparse} layer output, encoded into the vectors $\{\boldsymbol{y}_n\}_{n=1}^{N}\in\mathbb{R}^{K \cdot U}$ that are fed to the next layer. 
In the following, we encode the outcome of local competition between the units of each block via the discrete latent vectors $\boldsymbol{\xi}_n \in \mathrm{one\_hot(U)}^K$, where $\mathrm{one\_hot(U)}$ is an one-hot vector with $U$ components. These denote the winning unit out of the $U$ competitors in each of the $K$ blocks of the layer, when presented with the $n$th datapoint.

To allow for inferring which layer connections must be retained, we adopt concepts from the field of Bayesian nonparametrics. Specifically, we commence by introducing a matrix of binary latent variables, $\boldsymbol{Z} \in \{0,1\}^{J\times K}$. The $(j,k)$th entry therein is equal to one if the $j$th input is presented to the $k$th block, and equal to zero otherwise; in the latter case, the corresponding \emph{set of weights}, $\{w_{j,k,u}\}_{u=1}^U$, are effectively canceled out from the model. Subsequently, we impose an IBP prior over $\boldsymbol{Z}$, to allow for performing inference over it, in a way that 
promotes retention of the barely needed components, as explained in Section 2.1. Turning to the winner sampling procedure within each LWTA block, we postulate that the latent variables $\boldsymbol{\xi}_n$ are also driven from the layer input, and exploit the connection utility information encoded into the inferred $\boldsymbol{Z}$ matrices.

Let us begin with defining the expression of layer output, $\boldsymbol{y}_n \in \mathbb{R}^{K \cdot U}$.
Following the above-prescribed rationale, we have:
\begin{align}
[\boldsymbol{y}_n]_{ku} &= [\boldsymbol{\xi}_n]_{ku} \sum_{j=1}^J (w_{j,k,u} \cdot z_{j,k} ) \cdot [\boldsymbol{x}_{n}]_j \; \in \mathbb{R}
\end{align}
where we denote as $[\boldsymbol{h}]_l$ the $l$th component of a vector $\boldsymbol{h}$.
In this expression, we consider that the winner indicator latent vectors are drawn from a Categorical (posterior) distribution  of the form:
\begin{equation}
\label{eqn:dense_xi}
\scalemath{0.85}{
q([\boldsymbol{\xi}_n]_k) =  \mathrm{Discrete}\bigg([\boldsymbol{\xi}_n]_k \bigg| \mathrm{softmax}\big( \sum_{j=1}^J [w_{j,k,u}]_{u=1}^{U} \cdot z_{j,k} \cdot  [\boldsymbol{x}_{n}]_j \big) \bigg)
}
\end{equation}
where $[w_{j,k,u}]_{u=1}^{U}$ denotes the vector concatenation of the set $\{w_{j,k,u}\}_{u=1}^{U}$, and $[\boldsymbol{\xi}_n]_k \in \mathrm{one\_hot(U)}$.
On the other hand, the utility latent variables, $\boldsymbol{Z}$, are independently drawn from Bernoulli posteriors that read:
\begin{equation}
q(z_{j,k}) = \text{Bernoulli}(z_{j,k} | \tilde{\pi}_{j,k})
\end{equation}
where the $\tilde{\pi}_{j,k}$ are obtained through model training (Section 3.1).

Turning to the prior specification of the model latent variables, we consider a symmetric Discrete prior over the winner unit indicators, $[\boldsymbol{\xi}_n]_k \sim \mathrm{Discrete}(1/U)$, and an IBP prior over the utility indicators: 
\begin{equation}
\scalemath{0.9}{
u_k \sim \text{Beta}(\alpha,1) \quad \pi_k = \prod_{i=1}^k u_i \quad z_{j,k} \sim \text{Bernoulli}(\pi_k) \; \forall j.
}
\end{equation}

Finally, we define a distribution over the weight matrices, $\boldsymbol{W}$. To allow for simplicity, we impose a spherical prior $\boldsymbol{W} \sim \prod_{j,k,u} \mathcal{N}(w_{j,k,u}| 0, 1)$, and seek to infer a posterior distribution $q(\boldsymbol{W}) = \prod_{j,k,u} \mathcal{N}(w_{j,k,u} | \mu_{j,k,u},\sigma_{j,k,u}^2)$.
\begin{figure*}
\begin{center}
{\scalebox{.5}{\input{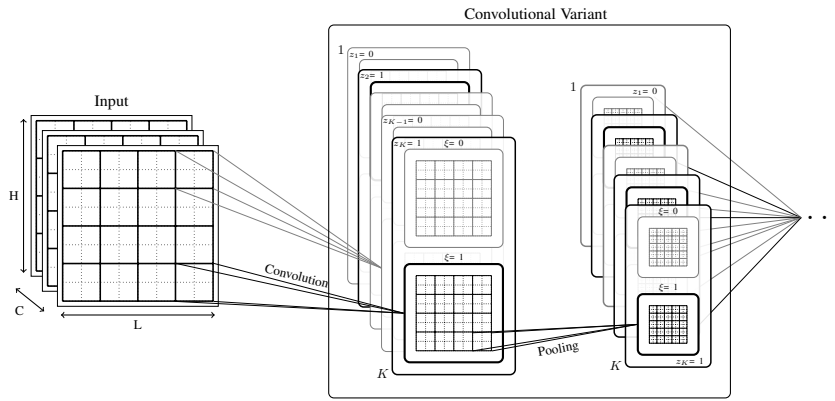}}}
\end{center}
\caption{A convolutional variant of our approach. Bold frames denote active (effective) kernels (LWTA blocks of competing feature maps), with $z=1$. Bold rectangles denote winner feature maps (with $\xi=1$).}
\label{fig:wta}
\end{figure*}

This concludes the formulation of a layer of the proposed SB-LWTA model. 

\subsection{A Convolutional Variant}

Further, we consider a variant of SB-LWTA which allows for accommodating convolutional operations. These are of importance when dealing with signals of 2D structure, e.g. images. To perform a convolution operation over an input tensor $\{\boldsymbol{X}\}_{n=1}^N \in \mathbb{R}^{H\times L \times C}$ at a network layer, we define a set of kernels, each with weights $\boldsymbol{W}_k\in \mathbb{R}^{h\times l \times C\times U}$, where $h,l,C,U$ are the kernel height, length, number of channels, and number of \emph{competing feature maps}, respectively, and $k \in \{1, \dots, K\}$. Hence, contrary to the grouping of linear units in LWTA blocks in Fig. 1, the proposed convolutional variant performs local competition among feature maps. That is, each (convolutional) kernel is treated as an LWTA block. Each layer of our convolutional SB-LWTA networks comprises multiple kernels of competing feature maps.  

We provide a graphical illustration of the proposed convolutional variant of SB-LWTA in Fig. 2. Under this model variant, we define the utility latent indicator variables, $z$, over whole kernels, that is full LWTA blocks. If the inferred posterior, $q(z_k=1)$, over the $k$th block is low, then the block is effectively omitted from the network. Our insights motivating this modeling selection concern the resulting computational complexity. Specifically, this formulation allows for completely removing kernels, thus \emph{reducing} the number of executed convolution operations. Hence, this construction facilitates efficiency, since convolution is computationally expensive.

Under this rationale, a layer of the proposed convolutional variant represents an input, $\boldsymbol{X}_n$, via an output tensor $\boldsymbol{Y}_n \in \mathbb{R}^{H\times L\times {K\cdot U}}$ obtained as the concatenation along the last dimension of the subtensors $\{[\boldsymbol{Y}_n]_{k}\}_{k=1}^{K} \in \mathbb{R}^{H\times L \times U}$ defined below:
\begin{align}
[\boldsymbol{Y}_n]_{k} &= [\boldsymbol{\xi}_n]_{k} \cdot \big( (\boldsymbol{W}_k \cdot z_{k}) \star \boldsymbol{X}_{n}\big)
\end{align}
where ``$\star$'' denotes the convolution operation and $[\boldsymbol{\xi}_n]_k \in \mathrm{one\_hot}(U)$. Local competition among \emph{feature maps within an LWTA block (kernel)} is implemented via a sampling procedure which is driven from the feature map output, yielding:
\begin{equation}
\scalemath{0.9}{
q([\boldsymbol{\xi}_n]_k) =  \mathrm{Discrete}\big([\boldsymbol{\xi}_n]_k \big| \mathrm{softmax} ( \sum_{h',l'} [ z_{k} \boldsymbol{W}_k  \star \boldsymbol{X}_{n} ]_{h',l',u} ) \big)
}
\end{equation}
We postulate a prior $[\boldsymbol{\xi}_n]_k \sim \mathrm{Discrete}(1/U)$. We consider
\begin{equation}
q(z_{k}) =\mathrm{Bernoulli}(z_{k}|\tilde{\pi}_k)
\end{equation}
with corresponding priors:
\begin{equation}
\scalemath{0.9}{
u_k \sim \text{Beta}(\alpha,1) \quad \pi_k = \prod_{i=1}^k u_i \quad z_{k} \sim \text{Bernoulli}(\pi_k)
}
\end{equation}
Finally, we again consider network weights imposed a spherical prior $\mathcal{N}(0,1)$, and seek to infer a posterior distribution of the form $\mathcal{N}(\mu,\sigma^2)$.

This concludes the formulation of the convolutional layers of the proposed SB-LWTA model. Obviously, this type of convolutional layer may be succeeded by a conventional pooling layer, as deemed needed in the application at hand.

\section{Training and Inference Algorithms}

\subsection{Model Training}

To train the proposed model, we resort to maximization of the resulting ELBO expression. Specifically, we adopt Stochastic Gradient Variational Bayes (SGVB) combined with: (i) the standard reparameterization trick for the postulated Gaussian weights, $\boldsymbol{W}$, (ii) the Gumbel-Softmax relaxation trick \cite{Maddison2017} for the introduced latent indicator variables, $\boldsymbol{\xi}$ and $\boldsymbol{Z}$; and (iii) the Kumaraswamy reparameterization trick \cite{Kumaraswamy1980} for the stick variables $\boldsymbol{u}$. 

Specifically, when it comes to the entailed Beta-distributed stick variables of the IBP prior, we can easily observe that these are not amenable to the reparameterization trick, in contrast to the postulated Gaussian weights. To address this issue, one can approximate their variational posteriors $q(u_k)=\mathrm{Beta}(u_k|a_k,b_k)$ via the Kumaraswamy distribution \cite{Kumaraswamy1980}:
\begin{align}
q(u_k;a_k,b_k) &= a_k b_k u_k^{a_k-1}(1-u_k^{a_k})^{b_k-1}
\end{align}
Samples from this distribution can be reparameterized as follows \cite{nali}:
\begin{align}
u_k = 
\left( 1 - (1-X)^{\frac{1}{b_k}}\right)^{\frac{1}{a_k}}, \; X \sim U(0,1)
\end{align}

On the other hand, in the case of the Discrete (Categorical or Bernoulli) latent variables of our model, performing back-propagation through reparameterized drawn samples becomes infeasible. Recently, the solution of introducing appropriate continuous relaxations has been proposed by different research teams \cite{Jang2017, Maddison2017}. Let $\boldsymbol{\eta} \in (0,\infty)^K$ be the \emph{unnormalized} probabilities of a considered Discrete distribution, $\boldsymbol{X}=[X_k]_{k=1}^{K}$, and $\lambda \in (0,\infty)$ be a hyperparameter referred to as the \textit{temperature} of the relaxation. Then, the drawn samples of $\boldsymbol{X}$ are expressed as differentiable functions of the form:
\begin{align}
\label{eqn:concrete_sampling}
X_k &= \frac{\exp(\log \eta_k+G_k)/\lambda)}{\sum_{i=1}^K \exp((\log \eta_i+G_i)/\lambda)}, \\ G_k &= -\log(-\log U_k), \; U_k \sim\text{Uniform}(0,1)
\end{align}
In our work, the values of $\lambda$ are annealed during training as suggested in \citet{Jang2017}.

We introduce the mean-field (posterior independence) assumption across layers, as well as among the latent variables $\boldsymbol{\xi}$ and $\boldsymbol{Z}$ pertaining to the same layer. All the posterior expectations in the ELBO are computed by drawing MC samples under the Normal, Gumbel-Softmax and Kumaraswamy repametrization tricks, respectively.  On this basis, ELBO maximization is performed using standard off-the-shelf, stochastic gradient techniques; specifically, we adopt ADAM \cite{kingma2014adam} with default settings.  For completeness sake, the expression of the eventually obtained ELBO that is optimized via ADAM is provided in the Supplementary Material. 

\subsection{Inference Algorithm}
Having trained the model posteriors, we can now use them to effect inference for unseen data. In this context, SB-LWTA offers two main advantages over conventional techniques: 

(i) By exploiting the inferred component utility latent indicator variables, we can naturally devise a method for omitting the contribution of components that are effectively deemed unnecessary. To this end, one may introduce a \emph{cut-off threshold}, $\tau$; any component with inferred corresponding posterior $q(z)$ below $\tau$ is omitted from computation. 

We emphasize that this mechanism is in stark contrast to recent related work in the field of BNNs; in these cases, utility is only \emph{implicitly inferred}, by \emph{thresholding higher-order moments} of hierarchical densities over the values of the \emph{network weights themselves}, $\boldsymbol{W}$ (see also related discussion in Sec. 1). For instance, \citet{Welling}  imposed the following prior over the network weights
\begin{equation}
z\sim p(z) \qquad w\sim\mathcal{N}(0,z^2)
\end{equation}
where $p(z)$ can be a Horseshoe-type or log-uniform prior. However, such a modeling scheme requires extensive heuristics for the appropriate, ad hoc, selection of the prior $p(z)$ hyperparameter values that can facilitate the desired sparsity, and the associated thresholds at each network layer. On the contrary, our principled paradigm enables fully automatic, data-driven inference of network utility, using \emph{dedicated} latent variables to infer which network components are needed. We only need to specify \emph{one global hyperparameter, that is the innovation hyperparameter $\alpha$  of the IBP, and one global truncation threshold}, $\tau$. Even more importantly, \emph{our model is not sensitive to small fluctuations of the values of these selections}. This is a unique advantage of our model compared to the alternatives, as it obviates the need of extensive heuristic search of hyperparameter values.

(ii) The provision of a full Gaussian posterior distribution over the network weights, $\boldsymbol{W}$, offers a natural way of reducing the floating-point bit precision level of the network implementation. Specifically, the posterior variance of the network weights constitutes a measure of uncertainty in their estimates. Therefore, we can leverage this uncertainty information to assess which bits are significant, and remove the ones which fluctuate too much under approximate posterior sampling. The unit round off necessary to represent the weights is computed by making use of the mean of the weight variances, in a fashion similar to \citet{Welling}. 

We emphasize that, contrary to \citet{Welling}, our model is endowed with the important benefit that the procedure of bit precision selection for the network weights \emph{relies on different posteriors} than the component omission process. We posit that by disentangling these two processes, we reduce the tendency of the model to underestimate posterior variance. Thus, we may yield stronger network compression while retaining predictive performance.

Finally, we turn to prediction generation. To be Bayesian, we need to sample several configurations of the weights in order to assess the predictive density, and perform averaging; this is inefficient for real-world testing. Here, we adopt a common approximation as in \citet{Welling,Neklyudov}; that is, we perform traditional forward propagation using the means of the weight posteriors in place of the weight values. Concerning winner selection, we compute the posteriors $q(\boldsymbol\xi)$ and select the unit with maximum probability as the winner; that is, we resort to a hard winner selection, instead of performing sampling. Lastly, we retain all network components the posteriors, $q(z)$, of which exceed the imposed truncation threshold, $\tau$.

\section{Experimental Evaluation}

In the following, we evaluate the two variants of our SB-LWTA approach. We assess the predictive performance of the model, and its requirements in terms of floating-point bit precision and number of trained parameters. We also compare the effectiveness of local competition among linear units to standard nonlinearities. 

\subsection{Implementation Details}

In our experiments, the stick variables are drawn from a $\mathrm{Beta}(1,1)$ prior. The hyperparameters of the approximate Kumaraswamy posteriors of the sticks are initialized as follows: the $a_k$'s are set equal to the number of LWTA blocks of their corresponding layer; the $b_k$'s are always set equal to $1$. All other initializations are random within the corresponding support sets. The employed cut-off threshold, $\tau$, is set to $10^{-2}$. The evaluated simple SB-LWTA networks omit \emph{connections} on the basis of the corresponding latent indicators $z$ being below the set threshold $\tau$. Analogously, when using the proposed convolutional SB-LWTA architecture, we omit full LWTA \emph{blocks} (convolutional kernels). 

\subsection{Experimental results} 

\begin{table}[h]
\vskip -0.1in
\caption{Classification accuracy and bit precision for the LeNet-300-100 architecture. All connections are retained. Bit precision refers to the necessary precision (in bits) required to represent the weights of each of the three layers.}
\vskip 0.1in
\label{table:activations}
\centering
{\scalebox{.9}{
\begin{small}
\begin{sc}
\begin{tabular}{lcc}
\toprule
Activation & Error($\%$) & Bit Precision (Error $\%$) \\
\midrule
ReLU    & 1.60 & 2/4/10 (1.62) \\
Maxout/2 units & 1.38 & 1/3/12 (1.57) \\
Maxout/4 units  & 1.67 & 2/5/12 (1.75) \\
SB-LWTA/2 units    & \textbf{1.31} & 1/3/11 (1.47)  \\
SB-LWTA/4 units     & 1.34 & \textbf{1/2/8} (1.5)\\
\bottomrule
\end{tabular}
\end{sc}
\end{small}
}}
\end{table}
\setlength{\textfloatsep}{10pt}

We first consider the classical LeNet-300-100 feedforward architecture. We initially assess LWTA nonlinearities regarding their classification performance and bit precision requirements, compared to ReLU and Maxout \cite{maxout} activations. To this end, we replace the $K$ LWTA blocks and the $U$ units therein (Fig. 1) with (i) $K$ maxout blocks, each comprising $U$ units, and (ii) $K\cdot U$ ReLU units (see supplementary material); no other regularization techniques are used, e.g., dropout. These alternatives are trained by imposing Gaussian priors over the network weights and inferring the corresponding posteriors via SGVB. We consider two alternative configurations comprising: 1)150 and 50 LWTA blocks on the first and second layer, respectively, of two competing units each; and 2) 75 and 25 LWTA blocks of four competing units. This experimental setup allows for us to examine the effect of the number of competing LWTA units on model performance, \emph{with all competitors initialized at the same number of trainable weights}.  We use MNIST in these experiments. 

\begin{table*}
\vskip -0.1in
\caption{Computational footprint reduction experiments. SB-ReLU denotes a variant of SB-LWTA using ReLU units.}
\vskip 0.1in
\centering
\label{table:pruning}
\begin{tabular}{ccccc}
\hline 
Architecture & Method & Error (\%) & \# Weights & Bit precision\\
\hline
\multirow{7}{*}{\shortstack{LeNet\\300-100}} & Original & 1.6 & $235K/30K/1K$  & $23/23/23 $ \Tstrut\\
& StructuredBP \cite{Neklyudov} & 1.7 & $23,664/6,120/450$ & $23/23/23 $\\
& Sparse-VD \cite{Molchanov} & 1.92 & $58,368/8,208/720$  & $8/11/14 $\\
& BC-GHS \cite{Welling} & 1.8 & $26,746/1,204/140$ & $13/11/10 $\\ 
& SB-ReLU & 1.75 & $13.698/6.510/730$ & $3/4/11 $\\
& SB-LWTA (2 units) & 1.7 & $12,522/6,114/534$ & $2/3/11 $\\ 
& SB-LWTA (4 units) & 1.75 & $23,328/9,348/618$ & $2/3/12 $\Bstrut\\
\hline 
\end{tabular}
\end{table*}

Further, we consider the LeNet-5-Caffe convolutional net, which we also evaluate on MNIST. The original LeNet-5-Caffe comprises 20 5x5 kernels (feature maps) on the first layer, 50 5x5 kernels (feature maps) on the second layer, and a dense layer with 500 units on the third. In our (convolutional) SB-LWTA implementation, we consider 10 5x5 kernels (LWTA blocks) with 2 competing feature maps each on the first layer, and 25 5x5 kernels with 2 competing feature maps each on the second layer. The intermediate pooling layers are similar with the reference architecture. We additionally consider an implementation comprising 4 competing feature maps deployed within 5 5x5 kernels on the first layer, and 12 5x5 kernels on the second layer, reducing the total feature maps of the second layer to 48. 

Finally, we perform experimental evaluations on a more challenging benchmark dataset, namely CIFAR-10 \cite{Krizhevsky}. To enable the wide replicability of our results within the community, we employ a computationally light convolutional architecture proposed by Alex Krizhevsky, which we dub ConvNet. The architecture comprises two layers with 64 5x5 kernels (feature maps), followed by two dense layers with 384 and 192 units respectively. Similar to LeNet-5-Caffe, our SB-LWTA implementation consists in splitting the original architecture into pairs of competing feature maps on each layer. For completeness sake, an extra experiment on CIFAR-10, dealing with a much larger network (VGG-like architecture), can be found in the provided Supplementary Material.

\textbf{LeNet-300-100.}
We train the network from scratch on the MNIST dataset, without using any data augmentation procedure. In Table \ref{table:activations}, we compare the classification performance of our approach, employing 2 or 4 competing LWTA units, to LeNet-300-100 configurations employing commonly used nonlinearities. The results reported in this Table pertaining to our approach, are obtained without omitting connections the utility posteriors, $q(z)$, of which fall below the cut-off threshold, $\tau$. In the second column of this Table, we observe that our SB-LWTA model offers competitive accuracy and improves over the considered alternatives when operating at full bit precision (float32). The third column of this Table shows how network performance changes when we attempt to reduce bit precision for both our model and the considered competitors\protect{\footnote{Following IEEE 754-2008,  floating-point data  representation comprises 3 different  quantities: a) $1$-bit sign, b) $w$ exponent bits. and c) $t=p-1$ precision in bits \cite{ieee754}. Thus, for the 32-bit format, we have $t=23$ as the original bit precision.}}. Bit precision reduction is based on the inferred weight posterior variance, similar to \citet{Welling} (see also the supplementary material).
As we observe, not only does our approach yield a clearly improved accuracy in this case, but it also imposes the lowest memory footprint.

The corresponding comparative results obtained when we employ the considered threshold to reduce the computational costs are depicted in Table \ref{table:pruning}. As we observe, our approach continues to yield competitive accuracy; this is on par with the best performing alternative, which requires, though, a significantly higher number of weights combined with up to an order of magnitude higher bit precision. Thus, our approach yields the same accuracy for a lighter computational footprint. Indeed, it is important to note that our approach remains at the top of the list in terms of the obtained accuracy while retaining the \emph{least number of weights}, despite the fact that it was initialized in the same dense fashion as the alternatives. Even more importantly, our method \emph{completely outperforms all the alternatives} when it comes to its final bit precision requirements. 

Finally, it is significant to note that by replacing in our model the LWTA blocks with ReLU units, a variant we dub SB-ReLU in Table \ref{table:pruning}, we yield clearly inferior outcomes. This constitutes strong evidence that LWTA mechanisms, at least the way implemented in our work, offer benefits over conventional nonlinearities.

\begin{table*}
\vskip -0.1in
\caption{Learned Convolutional Architectures.}
\vskip 0.1in
\label{table:conv}
\centering{}%
{\scalebox{.9}{
\begin{tabular}{ccccc}
\hline 
 Architecture & Method & Error (\%) & $\#$ Feature Maps (Conv. Layers) & Bit precision (All Layers)\Tstrut\\
\hline
\multirow{5}{*}{LeNet-5-Caffe} & Original & 0.9 & $20/50$  & $23/23/23/23$ \Tstrut\\
& StructuredBP \cite{Neklyudov} & 0.86 & $3/18$  & $23/23/23/23$\\
& VIBNet \cite{dai18d} & 1.0 & $7/25$ & $23/23/23/23$\\
& Sparse-VD \cite{Molchanov} & 1.0 & $14/19$ & $13/10/8/12$ \\
& BC-GHS \cite{Welling} & 1.0 & $5/10$ & $10/10/14/13$\\
& SB-ReLU & 0.9 & $10/16$ & $8/3/3/11$\\
& SB-LWTA-2 & 0.9 & $6/6$ & $6/3/3/13$\\
& SB-LWTA-4 & 0.8 & $8/12$ & $11/4/1/11$\Bstrut\\
\hline
\multirow{5}{*}{ConvNet} & Original & $17.0$ & $64/64$ & $23$ in all layers \Tstrut\Bstrut\\
& BC-GNJ\cite{Welling} & 18.6 & $54/49$ & $13/8/4/5/12$\Tstrut\Bstrut\\
& BC-GHS\cite{Welling} & 17.9 & $42/52$ & $12/8/5/6/10$\Tstrut\Bstrut\\
& SB-LWTA-2 & 17.5 & $40/42$ & $11/7/5/4/10$\Tstrut\Bstrut\\
\hline
\end{tabular}
}}
\end{table*}

\begin{figure}
\label{fig:most_activated}
\centering
\includegraphics[scale=0.32]{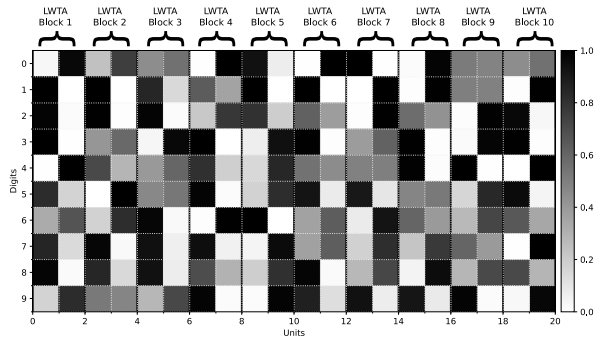}
\caption{Probabilities of winner selection for each digit in the test set for the first 10 blocks of the second layer of the LeNet-300-100 network, with two competing units; black denotes very high winning probability, while white denotes very low probability.}
\end{figure}

\textbf{LeNet-5-Caffe and ConvNet convolutional architectures.}
For the LeNet-5-Caffe architecture, we train the network from scratch. In Table 3, we provide the obtained  comparative effectiveness of our approach, employing 2 or 4 competing LWTA feature maps. Our approach requires the \textit{least number of feature maps} while at the same time offering significantly \textit{higher compression rates} in terms of bit precision, as well as better classification accuracy than the best considered alternative. By using the SB-ReLU variant of our approach, we once again yield inferior performance compared to SB-LWTA, reaffirming the benefits of LWTA mechanisms compared to conventional nonlinearities.

To obtain some comparative results, we additionally implement the BC-GNJ and BC-GHS models with the default parameters as described in \citet{Welling}. The learned architectures along with their classification accuracy and bit precision requirements are illustrated in Table 3. Similar to the LeNet-5-Caffe convolutional architecture, our method retains the \textit{least number of feature maps}, while at the same time provides \textit{the most competitive bit precision requirements} accompanied with \textit{higher predictive accuracy} compared to the competition.

\textbf{Further Insights.}
Finally, we scrutinize the competition patterns established within the LWTA blocks of an SB-LWTA network. To this end, we focus on the second layer of the LeNet-300-100 network with blocks comprising two competing units. Initially, we examine the distribution of the winner selection probabilities, and how they vary over the ten MNIST classes. In Figure 3, we depict these probabilities for the first ten blocks of the network, averaged over all the data points in the \emph{test set}. As we observe, the distribution of winner selection probabilities is unique for each digit. This provides empirical evidence that the trained winner selection mechanism successfully encodes salient discriminative  patterns with strong generalization value in the test set. Further, in Figure 4, we examine what the overlap of winner selection is among the MNIST digits. Specifically, for each digit, we compute the most often winning unit in each LWTA block, and derive the fraction of overlapping winning units over all blocks, for each pair of digits. It is apparent that winner overlap is quite low, typically below $50\%$; that is, considering any pair of digits, we yield an overlap in the winner selection procedure which is always below $50\%$. This is another strong empirical result reaffirming that the winner selection process encodes discriminative patterns of generalization value.

\textbf{Computational Times.} As a concluding note, let us now discuss the computational time required by SB-LWTA networks, and how it compares to the baselines. One training algorithm epoch takes on average 10\% more computational time for a network formulated under the SB-LWTA paradigm compared to a conventional network formulation (dubbed "Original" in Tables 2 and 3). On the other hand, prediction generation is immensely faster, since SB-LWTA significantly reduces the effective network size. For instance, in the LeNet-5-Caffe experiments, SB-LWTA reduces prediction time by one order of magnitude over the baseline.

\begin{figure}
\vskip -0.1in
\label{fig:pairs}
\centering
\includegraphics[scale=0.4]{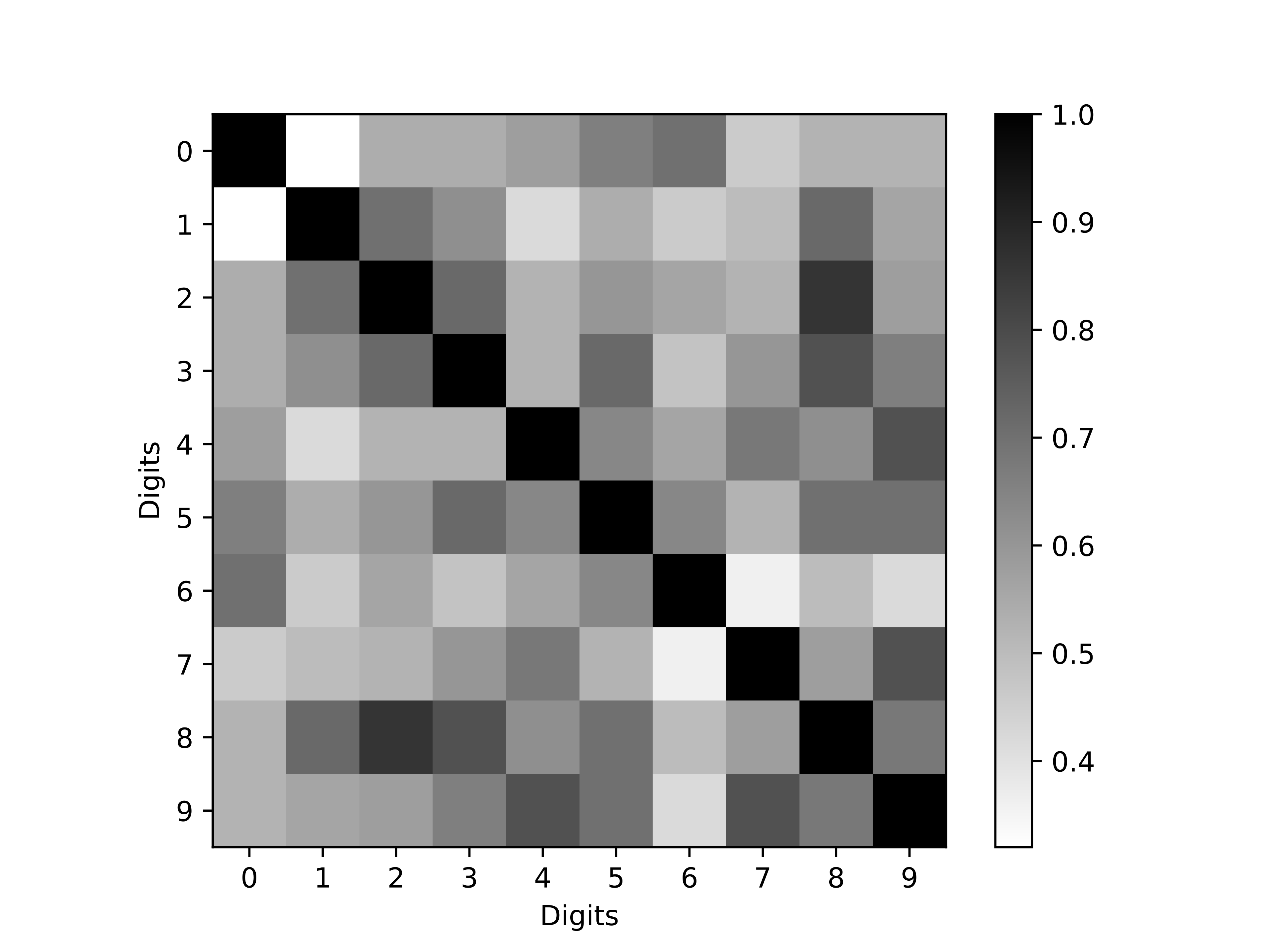}
\caption{MNIST dataset: Winning units overlap among digits. Black denotes that the winning units of all LWTA blocks are the same; moving towards white, overlap drops.}
\end{figure}
\setlength{\textfloatsep}{10pt}

\section{Conclusions}
In this paper, we examined how we can enable deep networks to infer, in a data driven fashion, the immensity of the computational footprint they need so as to effectively model a training dataset. To this end, we introduced a deep network principle with two core innovations: i) the utilization of LWTA nonlinearities, implemented as statistical arguments via discrete sampling techniques; ii) the establishment of a network component utility inference paradigm, implemented by resorting to nonparametric Bayesian processes. Our assumption was that the careful blend of these core innovations would allow for immensely reducing the computational footprint of the networks without undermining predictive accuracy. Our experiments have provided strong empirical support to our approach, which outperformed all related attempts, and yielded a state-of-the-art combination of accuracy and computational footprint. These findings motivate us to further examine the efficacy of these principles in the context of other challenging machine learning problems, including generative modeling and lifelong learning. These constitute our ongoing and future research work directions.

\section*{Acknowledgments}
We gratefully acknowledge the support of NVIDIA Corporation with the donation of the Titan Xp GPU used for this research. K. Panousis research was co-financed by Greece and the European Union (European Social Fund- ESF) through the Operational Programme ``Human Resources Development, Education and Lifelong Learning'' in the context of the project ``Strengthening Human Resources Research Potential via Doctorate Research'' (MIS-5000432), implemented by the State Scholarships Foundation (IKY). S. Chatzis research was partially supported by the Research Promotion Foundation of Cyprus, through the grant: INTERNATIONAL/USA/0118/0037.

\small
\bibliography{avi,bibl}
\bibliographystyle{icml2019}


\end{document}